\documentclass[letterpaper]{article} 
\usepackage{aaai2026}  
\usepackage{times}  
\usepackage{helvet}  
\usepackage{courier}  
\usepackage[hyphens]{url}  
\usepackage{graphicx} 
\usepackage{natbib}  
\usepackage{caption} 
\usepackage{algorithm}
\usepackage{algorithmic}
\usepackage{booktabs}
\usepackage{multirow}
\usepackage{amsmath}
\usepackage{tcolorbox}
\usepackage{newfloat}
\usepackage{listings}
\DeclareCaptionStyle{ruled}{labelfont=normalfont,labelsep=colon,strut=off} 
\floatstyle{ruled}
\newfloat{listing}{tb}{lst}{}
\floatname{listing}{Listing}
\title{ProgRAG: Hallucination-Resistant Progressive Retrieval and Reasoning over Knowledge Graphs}
\author{
    Minbae Park\equalcontrib\textsuperscript{\rm 1},
    Hyemin Yang\equalcontrib\textsuperscript{\rm 2},
    Jeonghyun Kim\textsuperscript{\rm 2},
    Kunsoo Park\textsuperscript{\rm 3},
    Hyunjoon Kim\thanks{Corresponding author.}\textsuperscript{\rm 1,2}
}
\affiliations{
    \textsuperscript{\rm 1}Department of Artificial Intelligence, Hanyang University, South Korea\\

    \textsuperscript{\rm 2}Department of Data Science, Hanyang University, South Korea\\
    \textsuperscript{\rm 3}Department of Computer Science and Engineering, Seoul National University, South Korea\\
    \textsuperscript{} \{pmb0323, hmym7308, gemma1126, hyunjoonkim\}@hanyang.ac.kr; kpark@thoery.snu.ac.kr
}

\usepackage{bibentry}

\begin{document}

\maketitle

\begin{abstract}
Large Language Models (LLMs) demonstrate strong reasoning capabilities but  struggle with hallucinations and limited transparency. Recently, KG-enhanced LLMs that integrate knowledge graphs (KGs) have been shown to improve reasoning performance, particularly for complex, knowledge-intensive tasks. However, these methods still face significant challenges, including inaccurate retrieval and reasoning failures, often exacerbated by long input contexts that obscure relevant information or by context constructions that struggle to capture the richer logical directions required by different question types. Furthermore, many of these approaches rely on LLMs to directly retrieve evidence from KGs, and to self-assess the sufficiency of this evidence, which often results in premature or incorrect reasoning. To address the retrieval and reasoning failures, we propose ProgRAG, a multi-hop knowledge graph question answering (KGQA) framework that decomposes complex questions into sub-questions, and progressively extends partial reasoning paths by answering each sub-question. At each step, external retrievers gather candidate evidence, which is then refined through uncertainty-aware pruning by the LLM. Finally, the context for LLM reasoning is optimized by organizing and rearranging the partial reasoning paths obtained from the sub-question answers. Experiments on three well-known datasets demonstrate that ProgRAG outperforms existing baselines in multi-hop KGQA, offering improved reliability and reasoning quality.
\end{abstract}

\begin{links}
    \link{Code}{https://github.com/hyemin-yang/ProgRAG}
\end{links}

\section{Introduction}
Since the emergence of large language models (LLMs) like ChatGPT, their remarkable reasoning abilities have demonstrated impressive performance in natural language processing tasks, particularly in question answering \cite{wei2022chain, brown2020language, wang2022self, besta2024graph,yao2023tree}. However, challenges such as hallucinations and limited performance on complex, knowledge-intensive tasks still persist \cite{ji2023survey}. These limitations have fueled growing interest in incorporating external structured knowledge sources, especially knowledge graphs (KGs), to enhance the reliability and accuracy of LLMs.

\begin{figure}[t]
\centering
\includegraphics[width=\columnwidth]{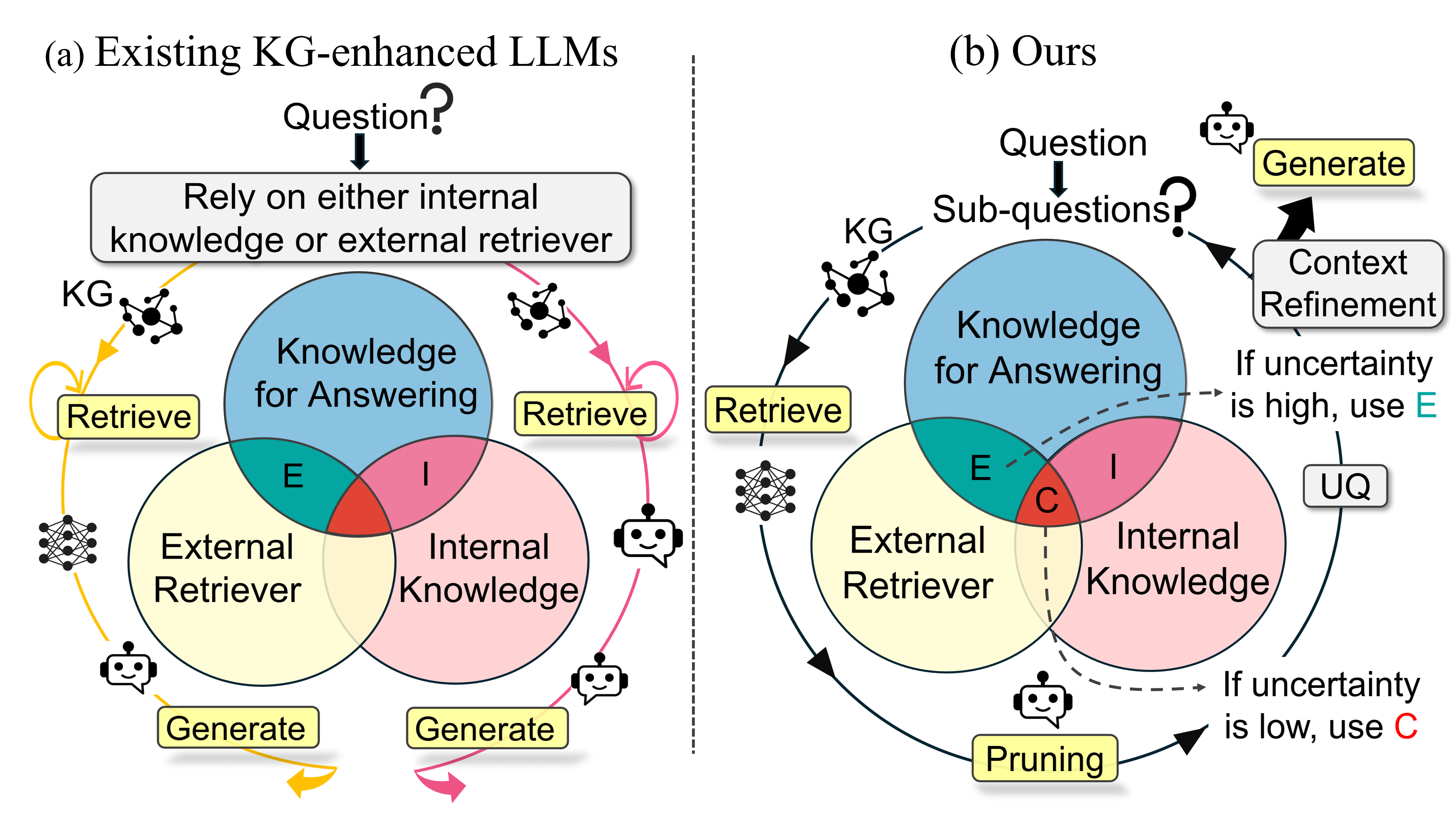}
\caption{Comparison between existing KG-enhanced LLMs and the proposed framework. E, I, and C denote External, Internal, and Core Supporting Evidences, respectively, and UQ in (b) indicates uncertainty quantification.}
\label{fig1}
\end{figure}

Recent advancements have led to the development of KG-enhanced LLMs \cite{pan2024unifying}, which integrate KGs in either the fine-tuning or inference phases. Despite these improvements, LLM fine-tuning methods \cite{mavromatis2024gnn, luo2023reasoning, luo2023chatkbqa} often require substantial computational resources and struggle to generalize effectively to unseen knowledge. A complementary line of methods using KGs only in the inference phase, i.e., KG-enhanced LLM inference methods, can be broadly categorized into two approaches in terms of retrieval, as illustrated in Figure \ref{fig1}(a): (1) LLM-as-retriever approach in which the LLM itself guides the retrieval process by leveraging its internal knowledge (corresponding to region I in the figure), and (2) external retriever-based approach in which an external retriever extracts relevant reasoning paths or subgraphs (corresponding to region E). Each approach relies on different sources of knowledge, i.e., internal or external, to retrieve supporting evidence for reasoning.

\begin{figure}[t]
\centering
\includegraphics[width=\columnwidth]{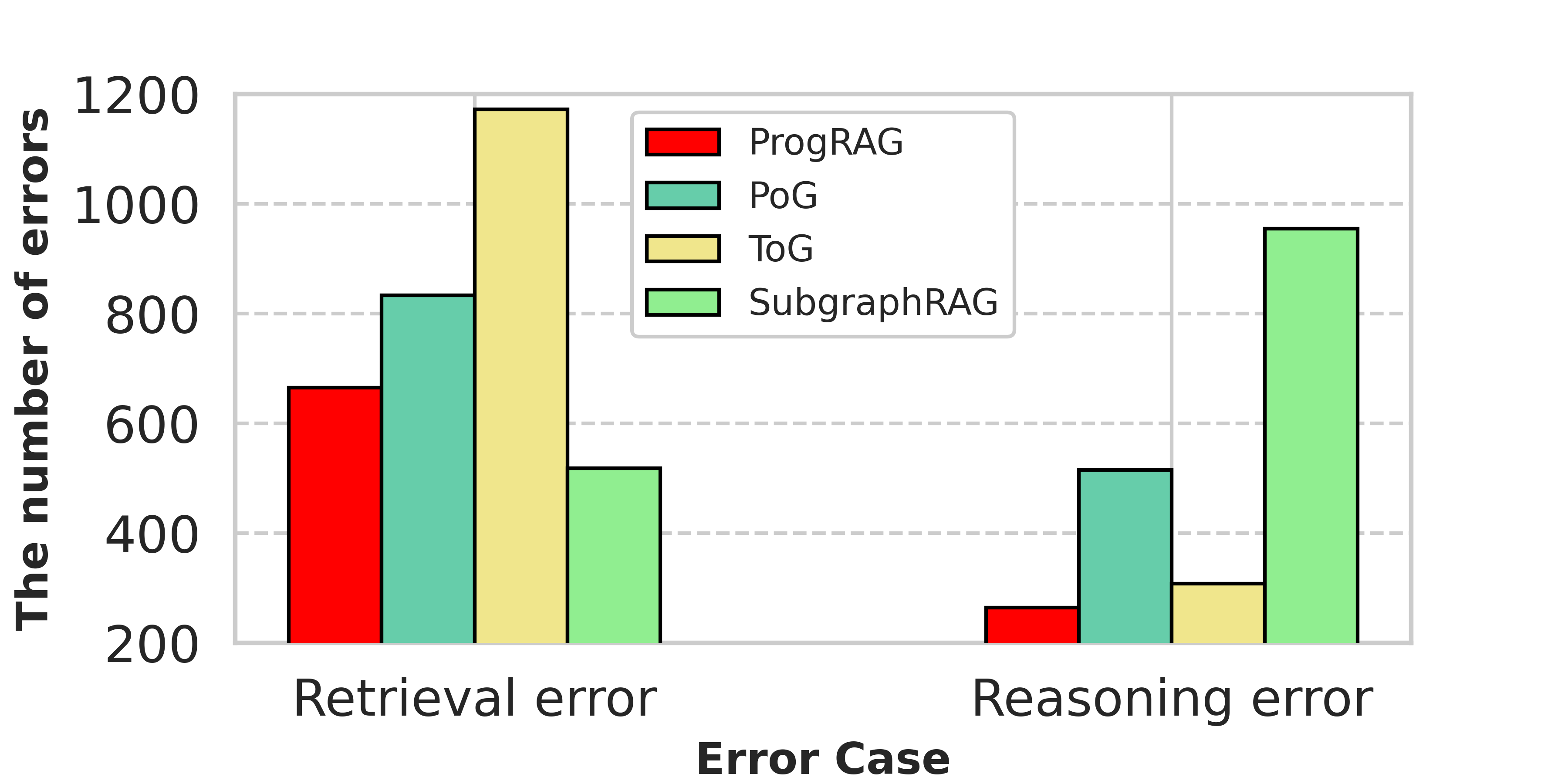}
\caption{Comparison of typical error cases in existing methods versus the proposed framework.}
\label{fig2}
\end{figure}

However, both approaches still face fundamental limitations. 
First, they often fail to retrieve accurate evidence. Second, even when correct supporting evidence is retrieved, the LLM frequently generates an incorrect answer. In Figure \ref{fig2}, we analyze representative failure cases on the CWQ dataset for LLM-as-retriever methods like ToG \cite{sun2023think} and PoG \cite{chen2024plan}, and the external retriever-based method SubgraphRAG \cite{li2024simple}. We define retrieval error as cases where the retrieved reasoning path or subgraph does not contain the answer entity, and reasoning error as cases where the answer is included in the retrieved evidence but the LLM fails to generate the correct answer. ToG and PoG exhibit high rates of both retrieval and reasoning errors. 

In contrast, SubgraphRAG extracts the top-100 question-relevant triples from the KG, thereby reducing retrieval error. Nonetheless, it still suffers from significant reasoning errors. These results can be attributed to four key factors: 
\begin{enumerate}
    \item Relying solely on the LLM to navigate the extensive search space of KGs is inherently ineffective, resulting in substantial retrieval errors.  
    \item The self-assessment used by LLM-as-retriever methods is prone to hallucination, often resulting in either premature termination or unnecessary continuation of the reasoning process.  
    \item Excessively long input contexts from numerous retrieved triples or paths dilutes relevant evidence and hinders answer identification~\cite{liu2023lost}. This affects SubgraphRAG, ToG, and PoG alike.
    \item Existing KG-enhanced LLMs have limited ability to handle richer logical structures beyond simple linear hops \cite{zhu2025beyond}, leading to suboptimal contexts for LLM reasoning, as later validated through our experiments.
\end{enumerate}

To address the challenges of retrieval and reasoning errors particularly those arising from the hallucinatory behavior of LLMs, we propose a progressive retrieval and reasoning framework, \textbf{ProgRAG}, as illustrated in Figure \ref{fig1}(b): \textbf{(i)} we decompose a complex question into sequential sub-questions, and answer each one iteratively, thereby progressively extending reasoning paths. For this, we treat the number of sub-questions as the exploration depth of its complete reasoning path, adapting it to balance sufficient reasoning with avoidance of over-exploration;
 \textbf{(ii)} for each sub-question, our external retrievers produces relevant evidence supporting the answer for that sub-question from the KG, which is then filtered by the LLM to improve precision. This retrieval-and-pruning process narrows the search space, allowing the LLM to focus on a more relevant candidate set. 
Moreover, we adaptively increase reliance on external knowledge when the uncertainty of the LLM’s predictions is high; \textbf{(iii)} to reduce hallucination during reasoning, we refine and enhance the LLM’s input context using effective prompting techniques such as prefix enumeration and repacking, which structure the input to better emphasize relevant evidence, thereby improving the model’s ability to generate accurate and grounded answers. We evaluate ProgRAG on three KGQA benchmarks—WebQSP, CWQ, and CR-LT—and observe state-of-the-art performance on all datasets. ProgRAG outperforms the best baseline by 3.3\% on WebQSP, 4.9\% on CWQ, and 10.9\% on CR-LT in accuracy. 

\begin{itemize}
    \item We propose ProgRAG, a novel progressive retrieval and reasoning framework for multi-hop KGQA that iterates external retrieval and LLM-based pruning to address both retrieval and reasoning errors. 
    \item We first propose to refine the LLM context with various reasoning paths, enabling ProgRAG to adaptively handle different question types, improving the robustness and generalization of reasoning.
    \item We conduct comprehensive experiments on three multi-hop KGQA datasets, demonstrating that ProgRAG achieves state-of-the-art performance even with smaller LLMs, e.g., Gemma2-9B-it, GPT-4o-mini, without fine-tuning.
\end{itemize}

\section{Related Work}

Most KG-enhanced LLMs can be categorized into two approaches based on how LLMs are utilized. Several methods \cite{mavromatis2024gnn, long2025eperm, luo2024graph, luo2023chatkbqa, xu2025harnessing, luo2023reasoning, liu2025symagent, ao2025lightprof} incorporate KGs during the fine-tuning of LLMs. 

However, these approaches often generalize poorly to unseen knowledge and incur significant computational cost \cite{baek2023knowledge, axelsson2023using, he2024g}. An alternative line of work retrieves structured evidence from KGs and feeds it into LLMs for reasoning without fine-tuning, using either LLM-internal knowledge or external retrievers.

\textbf{LLM-as-retriever methods} rely on the LLMs to directly explore the KG and iteratively accumulate requisite information. StructGPT \cite{jiang2023structgpt} generates executable SQL queries to extract relevant KG evidence. ToG \cite{sun2023think} explores multiple reasoning paths within a predefined exploration breadth by retrieving query-relevant relations, followed by the corresponding entities. PoG \cite{chen2024plan} decomposes the question into sub-tasks and prompts all sub-tasks, along with the question, to the LLM to guide answer prediction. Building on ToG’s framework, ReKnoS \cite{wang2025reasoning} and MFC \cite{zhang2025good} further extend the search space using super-relations or meta-entities. However, due to unfiltered exploration, these methods often retrieve suboptimal paths and exhibit hallucination tendencies \cite{liu2023lost, dhole2025retrieve}. 

\textbf{External retriever-based methods} adopt a pipeline in which fine-tuned external retrieval models efficiently extract relevant subgraphs or reasoning paths from the KG \cite{jiang2022unikgqa, jiang2023reasoninglm, liu2024explore}. SubgraphRAG \cite{li2024simple} integrates a lightweight MLP with a parallel triple-scoring mechanism for subgraph retrieval. KG-CoT \cite{zhao2024kg} explores KG stepwise from question entities, generates reasoning paths via transition matrices, However, these methods either retrieve excessive structural information or fail to leverage LLMs’ internal knowledge during retrieval, resulting in low reasoning accuracy due to both irrelevant volume and insufficient precision.

\section{Preliminaries}

\textbf{Knowledge Graph (KG).} A knowledge graph is a set of factual triples, denoted by $\mathcal G=\{(e, r, e')|e,e'\in \mathcal E, r \in \mathcal R\}$, where $\mathcal E$ is a set of entities and $\mathcal R$ is a set of relations. Each triple $(e, r, e')$ represents a fact indicating that the head entity $e$ is connected to the tail entity $e'$ via relation $r$. 

\noindent\textbf{Knowledge Graph Question Answering (KGQA).} Given a natural language question $q$, a knowledge graph $\mathcal G$, and a set of key entities $\mathcal E_q \subseteq \mathcal E$ extracted from $q$, the goal of KGQA is to predict the set $\mathcal A_q \subseteq \mathcal E$ of answers to the question  $q$ by performing reasoning over $\mathcal G$.

\noindent\textbf{Reasoning Path.} A reasoning path is a sequence of consecutive triples in a KG, denoted as $(e_0, r_1,e_1)\rightarrow(e_1, r_2, e_2)\rightarrow ...\rightarrow (e_{d-1}, r_d, e_d)$, where each triple $(e_{i-1}, r_i, e_i) \in \mathcal{G}$. This sequence forms the reasoning trajectory from a key entity $e_0$ toward predicting the answer entity $e_d$. A prefix of a reasoning path is a subsequence of consecutive triples taken from the beginning of the path. It consists of the first $k$ triples of the full reasoning path where $k\le d$. 

\section{Method}

\subsection{Overview}
Figure \ref{fig:model} illustrates the ProgRAG framework, which performs three stages. First, ProgRAG identifies key entities from a given question, and initializes a partial reasoning path as each key entity, and decomposes the question into multiple sub-questions in the question decomposition stage. Second, in the sub-question answering stage, ProgRAG iteratively answers each sub-question by extending the partial reasoning paths obtained from the previous iteration. Finally, in the prefix enumeration and repacking stage, ProgRAG reorganizes all complete reasoning paths and their partial reasoning paths to form a structured context. The LLM then infers the final answer based on this context.

\subsection{Question Decomposition}

In this stage, ProgRAG decomposes a question into simpler sub-questions. Specifically, we first identify key entities, i.e., the entities central to the semantics of the question. 
The LLM decomposes the question into sub-questions and associates each one with its corresponding key entity, which is called key entity mapping.
For every key entity $e_s$, the LLM further decomposes its initial sub-question into more granular ones if possible; retains this sub-question as atomic otherwise. The full prompt is provided in Appendix~K. 
Consequently, we obtain a chain $Q_{e_s} = \{q_1, ..., q_d\}$ of sub-questions for every key entity $e_s$, where the depth $d$ represents the number of reasoning steps, i.e., iterations, in the subsequent stage. In this manner, ProgRAG can dynamically adjust the exploration depth depending on the question complexity, enabling more flexible and precise reasoning. 
Previous work such as Chain-of-Question \cite{yixing2024chain}  also decomposes a complex question. Unlike our method, this decomposition does not take key entities into account.

\subsection{Sub-question Answering}

\begin{figure*}[ht]
\centering
\includegraphics[width=\linewidth]{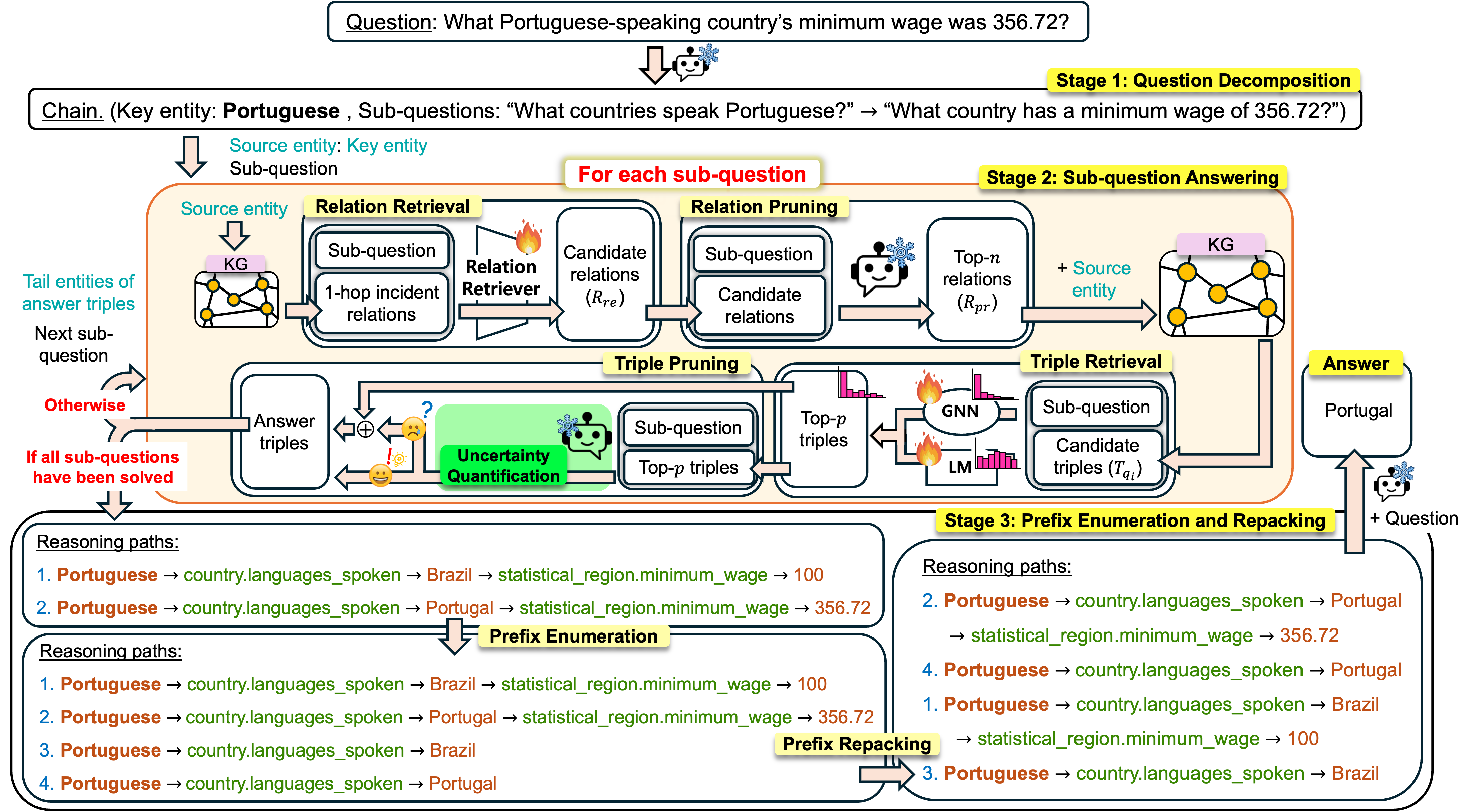}
\caption{ProgRAG operates in three stages: (1) Question decomposition, where the question is split into sub-questions based on a key entity; (2) Sub-question answering, where partial reasoning paths are progressively extended through retrieval and pruning; and (3) Prefix enumeration and repacking, where all prefixes of the reasoning paths are enumerated and reordered. Finally, the LLM infers the answer based on these prefixes.}
\label{fig:model}
\end{figure*}

For simplicity, we focus on a question with a single key entity, but our method can be easily extended to a question with multiple key entities. From the key entity and its chain of sub-questions, the sub-question answering stage iteratively finds the answer to each sub-question by extending the partial reasoning paths one hop at a time. At the first iteration, the source entity for each sub-question is the key entity; in subsequent iterations, the source entity is the answer entity from the previous iteration. At each iteration $i$, ProgRAG takes the source entity and sub-question $q_i$ as input, and performs four procedures: (1) relation retrieval, (2) relation pruning, (3) triple retrieval, and (4) triple pruning, as shown in Figure \ref{fig:model}\footnote{For questions with multiple key entities, their corresponding sub-question chains are solved independently, and the resulting reasoning paths are fed into the next stage.}.

\subsubsection{Relation Retrieval}\label{sec:rel_retrieve}

Seminal prior work in KGQA struggles with retrieving relevant relations from the KG, despite its significant impact on question answering quality. On the one hand, some methods pass a large number of relations linked to the final entities of partial reasoning paths to the LLM \cite{sun2023think}, leading to context overload. On the other hand, other methods retrieve relevant relations from the entire KG \cite{zhao2024kg}, often neglecting the local context needed for step-wise reasoning. 

For progressive relation retrieval, we select candidate relations $R_{re}(q_i)$ from the KG that are contextually relevant to a given sub-question $q_i$. To address the aforementioned limitations, we utilize a SentenceBERT-based cross-encoder \cite{reimers2019sentence, reimers2020curse} to score and rank one-hop triples $(s_i, r, e') \in G$ connected to the source entity $s_i$ based on their semantic relevance to $q_i$. The top-$m$ unique relations with the highest scores are selected as $R_{re}(q_i)$. 

\subsubsection{Relation Pruning}

For source entity $s_i$, sub-question $q_i$, and $R_{re}(q_i)$, the LLM selects the top-$n$ ($n\leq m$) relevant relations $R_{pr}(q_i)\subseteq R_{re}(q_i)$. All relations in $R_{re}(q_i)$ are sorted in descending order of their semantic similarity to $q_i$ in this prompt. To encourage the LLM to articulate its step-by-step reasoning, we apply Chain-of-Thought prompting (CoT) \cite{wei2022chain} leveraging our key observations on the Freebase KG widely used in this study: this KG organizes each relation into a three-level hierarchy, i.e., \textit{domain.source\_type.target\_type}. This hierarchy captures the semantic structure of the relation and its alignment with the associated triple, e.g., the relation of the triple \textit{(``Lou Seal", ``sports.mascot.team", ``San Francisco Giants")} consists of domain \textit{sports}, source type \textit{mascot}, and target type \textit{team}. 

Using CoT, the LLM infers three elements step by step: (1) the type of source (e.g., \textit{mascot}), (2) the target type (e.g., \textit{team}), and (3) a relational phrase that links them semantically (e.g., “mascot for”). We observe that each sub-question typically corresponds to a single-hop triple, making the step-by-step reasoning process highly compatible with the question decomposition.

In contrast to the prior work \cite{sun2023think} where the LLM directly evaluates the relevance of all retrieved relations to the question without the structure-aware step-by-step reasoning, we leverage hierarchical relation structure to enable the LLM to identify the relevant relations through reasoning. This process enhances interpretability and consistency.\footnote{Although this prompt is designed for hierarchical relations, it also performs effectively on CR-LT, a Wikidata-based dataset without hierarchical relation structure, where we observe consistently strong performance.} The full prompt is provided in Appendix~K. 

\subsubsection{Triple Retrieval}
As noted earlier, excessively long contexts can impair the focus of LLM on relevant information, increasing the risk of hallucinations and retrieval errors \cite{liu2023lost, liu2025towards}. To construct a more compact context, we prune unpromising candidate triples unlikely to lead to the answer for $q_i$ based on their relevance to $q_i$. The set $T_{q_i}$ of candidate triples consists of all triples that include the source entity $s_i$ and any relation in $R_{pr}(q_i)$. The relevance score between a triple and $q_i$ is computed by summing two components: (1) the textual semantic-based triple score, and (2) the structure-based entity score. 

\subsubsection{Textual Semantic-based Triple Scoring}
We employ a bi-encoder architecture composed of two MPNet-base models~\cite{song2020mpnet}, each generating embeddings for $q_i$ and for every $t\in T_{q_i}$, respectively. The textual semantic-based triple score $\Phi_t$ is computed by applying a softmax to the cosine similarity between the sub-question embedding $\mathbf{x}_{q_i}$ and the triple embedding $\mathbf{x}_t$, capturing their textual semantic relevance:
\begin{equation}\Phi_t(q_i)=\text{softmax}(\{\text{cos}(\mathbf{x}_{q_i}, \mathbf{x}_t)\}_{t\in T_{q_i}})\end{equation}

The encoders are fine-tuned using the InfoNCE~\cite{oord2018representation} loss to learn semantic alignment between sub-questions and their corresponding golden triples.

\subsubsection{Structure-based Entity Scoring}

In KGs, relations carry rich contextual information, making relation-centric approaches crucial for reasoning tasks \cite{wang2021relational}. To leverage this, we employ a query-dependent graph neural network (GNN) \cite{luo2025gfm}, which is designed to focus on the semantic relevance of relations in the context of specific queries. We embed all relations in the graph and the sub-question $q_i$ using the pretrained text encoder \cite{li2023towards}. Each entity is initialized with a zero vector, with the exception of the source entity $s_i$, which is assigned the sub-question embedding.

After passing through all GNN layers, we obtain the final representation for each entity in the graph. Let $E_{q_i}$ be the set of all tail entities in $T_{q_i}$. The representations of all entities in $E_{q_i}$ are concatenated and passed through an MLP followed by a softmax function, producing a probability that each $e\in E_{q_i}$ is the answer entity for $q_i$:
\begin{equation}
\Phi_e(q_i) = \text{softmax}(\text{MLP}(\text{concat}(\{\mathbf{h}_e\}_{e\in E_{q_i}})))
\end{equation}
For each $t\in T_{q_i}$, we compute the probability $u_t$ that triple $t$ is the answer triple for $q_i$:
\begin{equation}
u_t = \text{softmax}(\Phi_t(q_i) + \Phi_e(q_i))
\end{equation}
The above two components offer complementary perspectives, i.e., semantic relevance and structural context, enhancing both the accuracy of triple ranking. Finally, top-$p$ sampling \cite{holtzman2019curious} is applied over the probabilities $u_t$, sequentially selecting triples in descending order of probability until the cumulative score exceeds a predefined threshold $p$.

\subsubsection{Triple Pruning}

For the sub-question $q_i$, let $T_{re}(q_i)$ denote the set of triples remaining after top-$p$ sampling. From $T_{re}(q_i)$, we aim to obtain answer triples, with the source entity $s_i$ as the head and the predicted answer to $q_i$ as the tail.
Specifically, all triples $t\in T_{re}(q_i)$ are sorted in descending order of $u_t$ and provided to the LLM along with $q_i$.\footnote{If the LLM identifies plausible answers in $T_{re}(q_i)$, it then outputs them; otherwise, it returns “None”, indicating insufficient evidence. In such cases, ProgRAG returns to the relation pruning step with the unused relations in $R_{re}(q_i) \setminus R_{pr}(q_i)$ to explore alternative paths, thereby enabling early termination of incorrect reasoning and reducing redundant inference.} ProgRAG then assesses the reliability of the LLM outputs to detect its potential hallucinations, preventing error propagation in multi-hop reasoning, which is called ``Uncertainty Quantification''.

ProgRAG measures the aleatoric uncertainty (AU) of the LLM outputs by applying evidential modeling \cite{sensoy2018evidential,ma2025estimating} to the top-$K$ logits at the generation step in which the LLM produces the first token of the answer entity. If the uncertainty exceeds a predefined threshold, indicating low model confidence, the response is refined with the top-$l$  triples from $T_{re}(q_i)$ as external evidence. Otherwise, the response is accepted as is. If $q_i$ is the final sub-question, the model proceeds to the subsequent stage; otherwise, it continues to the next sub-question $q_{i+1}$. 




\subsection{Prefix Enumeration and Repacking}
In the final stage, the LLM infers the answer to the question $q$ based on all reasoning paths explored thus far. To enhance its reasoning ability, we enumerate all prefixes of every reasoning path, which are then included in the LLM context, as shown in Stage 3 of Figure~\ref{fig:model}. Next, we rank the prefixes by their semantic relevance to $q$, as scored by the triple retriever. Finally, the reordered prefixes and the question $q$ are fed into the LLM, guiding it to focus on the most relevant evidence and improving the accuracy of its answer. By carefully manipulating partial reasoning paths, the LLM reasons from prefixes that capture intermediate states, better aligning even partial paths with the nuanced semantics of the question, particularly when answers lie in the middle of reasoning paths or when multiple constraints must be satisfied. 

\section{Experiments}

To demonstrate the effectiveness and efficiency of ProgRAG on multi-hop KGQA, we conduct comprehensive experiments to address four research questions;
(\textbf{RQ1}) How effective is ProgRAG in the multi-hop reasoning for the KGQA task?;
(\textbf{RQ2})  How does each component in ProgRAG contribute to the overall performance?;
(\textbf{RQ3}) How accurately does ProgRAG retrieve ground-truth reasoning paths?;
(\textbf{RQ4}) How efficiently does ProgRAG perform multi-hop reasoning?

\subsection{Experimental Setup}

We adopt three publicly available multi-hop KGQA datasets: WebQuestionsSP (WebQSP) \cite{yih2016value}, ComplexWebQuestions (CWQ) \cite{talmor2018web}, and CR-LT-KGQA \cite{guo2024cr}. Dataset statistics and details are provided in Appendix~A.

As per prior works \cite{sun2023think, chen2024plan}, we use exact match accuracy (Hit@1) as our primary evaluation metric.
We use two small-scale LLMs for reasoning, i.e., Gemma 2-9b-it and GPT-4o-mini. Since black-box LLMs like GPT-4o-mini do not provide access to their logits, uncertainty quantification is only applied when using Gemma 2-9b, which is the main LLM for most experiments.
Other implementation details are further documented in the Appendix~B.


\subsection{Baseline Methods}
We evaluate the performance of ProgRAG against a diverse set of baselines spanning four major categories: (a) LLM-only prompting methods \cite{brown2020language, wei2022chain, wang2022self}; (b) Fine-tuned LLM-based methods \cite{luo2023reasoning, liu2025symagent, yu2022decaf, ao2025lightprof}; (c) LLM-as-retriever methods \cite{sun2023think, chen2024plan, jiang2023structgpt, wang2025reasoning, zhang2025good, liang2025fast}; (d) External retriever-based methods \cite{zhao2024kg, li2024simple}.\footnote{We do not extensively compare with fine-tuning-based \cite{mavromatis2024gnn} or semantic parsing approaches \cite{luo2023chatkbqa, xu2025harnessing}, as they follow fundamentally different paradigms from our retrieval-augmented setting.} 

\begin{table}[t]
\centering
\setlength{\tabcolsep}{1.6mm}{
\begin{tabular}{lccc}
\toprule
\multicolumn{1}{c}{\multirow{2}{*}{Method}} & \multirow{2}{*}{Category} & WebQSP & CWQ \\  \cmidrule(lr){3-4}

\multicolumn{1}{c}{} &   & Hit@1   & Hit@1   \\ \midrule
Zero-shot (GPT-3.5)& a & 54.4 & 34.9 \\
Few-shot (GPT-3.5)& a & 56.3 & 38.5 \\
CoT (GPT-3.5)& a & 57.4 & 43.2 \\ \midrule
\multicolumn{4}{c}{Fine-tuned LLM}         \\ \midrule
DeCAF (FiD-3B)  & b & 82.1 & \underline{70.4} \\
RoG (Llama2-7B-Chat)  & b & \underline{85.7} & 62.6 \\
SymAgent (Qwen2-7B)   & b & 78.5 & 58.9 \\
LightPROF (Llama3-8B)   & b & 83.8 & 59.3 \\\midrule
\multicolumn{4}{c}{Open source LLM or GPT-3.5} \\ \midrule
StructGPT (GPT-3.5)   & c & 75.2 & 55.2 \\
ToG (GPT-3.5)         & c & 76.2 & 58.9 \\
PoG (GPT-3.5)         & c & 82.0 & 63.2 \\
ReKnoS (GPT-3.5)      & c & 81.1 & 58.5 \\
MFC (GPT-3.5)         & c & 78.9 & 62.8 \\
KG-CoT (GPT-3.5)      & d & 82.1 & 51.6 \\
SubgraphRAG (GPT-3.5) & d & 83.1 & 56.3 \\
ProgRAG (Gemma2-9b)   &  & \textbf{88.5} & \textbf{73.7} \\ \midrule
\multicolumn{4}{c}{GPT-4o-mini}        \\ \midrule
StructGPT            & c & 79.5 & 64.7 \\
ToG                  & c & 77.0 & 59.0 \\
PoG                  & c & 83.2 & 63.5 \\
ReKnoS               & c & 83.8 & \underline{68.8} \\
MFC                  & c & 79.1 & 63.4    \\
FastToG              & c & 65.8 & 45.0 \\
KG-CoT               & d & OOM  & OOM  \\
SubgraphRAG          & d & \underline{86.2} & 58.3 \\
ProgRAG*       &   & \textbf{90.4} & \textbf{73.3} \\ \bottomrule
\end{tabular}
}
\caption{Performance comparison on WebQSP and CWQ. Baselines are categorized into (a) LLM-only prompting methods, (b) methods using fine-tuned LLMs, (c) LLM-as-retriever methods, and (d) external retriever-based methods. ProgRAG* denotes ProgRAG without Uncertainty Quantification, due to the black-box nature of GPT.}
\label{tab:performance}
\end{table}

\begin{table}[h]
\centering
\begin{tabular}{lc}
\toprule
          Method    & Hit@1 \\ \midrule
ToG (GPT-4o-mini)   & 50.0  \\
PoG (GPT-4o-mini)   & 56.6  \\
MFC (GPT-4o-mini)   & \underline{61.7}  \\
ProgRAG (Gemma2-9b) & \textbf{68.4}  \\ \bottomrule
\end{tabular}
\caption{Performance comparison on the CR-LT dataset.}
\label{tab:crlt_performance}
\end{table}

\subsection{RQ1: Main Results}
As shown in Table~\ref{tab:performance} and Table~\ref{tab:crlt_performance}, ProgRAG outperforms all baselines on WebQSP, CWQ, and CR-LT. 
Even without fine-tuning, ProgRAG with Gemma2-9b surpasses fine-tuned LLMs, outperforming RoG by 2.8\% on WebQSP and 11.1\% on CWQ. Compared to LLM-as-retriever methods, ProgRAG shows average improvements of 16.1\% on WebQSP and 29.7\% on CWQ, demonstrating the effectiveness of combining external retrieval with LLM-based pruning. ProgRAG outperforms external retriever-based methods by 7.1\% on WebQSP and 17.0\% on CWQ, with the large gain on CWQ highlighting the benefits of its progressive reasoning strategy for complex multi-hop queries requiring greater reasoning depth. Hop-wise performance details are provided in Appendix~H.
On CR-LT, ProgRAG outperforms all baselines, demonstrating its effectiveness in handling more complex queries.

\subsection{RQ2: Ablation Study}
\subsubsection{Effectiveness of Individual Techniques} 
Table \ref{tab:ablation} reports the performance of ProgRAG and its ablated variants. The definition of each variant is provided in Appendix~D.
All variants show performance degradation, indicating that each component contributes positively to the overall system. On WebQSP, the most significant drops occur when relation pruning (3.8\%) and triple retrieval (1.9\%) are removed. These declines are even larger on CWQ, highlighting the importance of combining external retrieval with LLM-based filtering to accurately identify supporting evidence. 
The largest drop of 22.6\% occurs when question decomposition is removed on CWQ, highlighting the importance of progressive reasoning through sub-question chains for complex queries. This suggests that using the number of sub-questions as exploration depth, rather than LLM self-assessment to decide whether to continue or stop exploration, helps avoid hallucinations and ensures more stable depth control. The second largest performance degradation in CWQ occurs when prefix enumeration is omitted, indicating that explicitly presenting diverse partial reasoning paths is essential to validate multiple contraints for complex multi-hop queries\footnote{The impact of ``prefix enumeration" and ``key entity mapping" on WebQSP is minimal due to its predominance of 1-hop or simple composition questions.}. Experimental results comparing with existing question decomposition methods \cite{zhang2025good} are presented in Appendix~J. 

\begin{table}[t]
\centering
\begin{tabular}{lccc}
\toprule
Method    &  WebQSP        & CWQ \\ \midrule
ProgRAG                 & \textbf{88.5}          & \textbf{73.7}        \\
w/o Prefix Enumeration     & \textbf{88.5}          & 63.9        \\
w/o Relation Pruning       & 84.2          & 65.4        \\
w/o Triple Retrieval       & 86.1          & 67.5        \\
w/o Uncertainty Quantification  & 87.2          & 68.5        \\
w/o Relation Retrieval     & 86.4          & 70.1        \\
w/o Prefix Repacking              & 86.9          & 70.2        \\
w/o Triple Pruning         & 87.8          & 72.8        \\
w/o Question Decomposition & 88.0          & 51.1        \\ 
w/o Key Entity Mapping     & \textbf{88.5}          & 70.0        \\ 
\bottomrule
\end{tabular}
\caption{Ablation study of the proposed methods.}
\label{tab:ablation}
\end{table}

\subsubsection{Performance Analysis by Question Type}
Table \ref{tab:question_types} presents the Hit@1 scores of ProgRAG and several state-of-the-art or well-known KG-enhanced LLMs—ToG, PoG, MFC, and SubgraphRAG—across different CWQ question types. 
ProgRAG consistently outperforms all the baselines across all question types. Notably, for superlative and comparative questions, where answers often reside at intermediate nodes of reasoning paths and thus require more structured reasoning, ProgRAG achieves scores of 70.6 and 75.6, respectively, while other methods show fall behind limited effectiveness. For conjunction questions, which demand satisfying multiple constraints across distinct paths, ProgRAG attains a Hit@1 of 76.6, substantially outperforming the baselines. 
The robustness of ProgRAG stems from prefix enumeration and repacking, which guide the LLM to focus on relevant partial reasoning paths and verify the constraints required by the question, leading to more accurate reasoning. Removing either component leads to substantial performance drops, particularly on complex questions\footnote{A slight decrease is observed for composition questions, where full-path reasoning is critical and partial path prioritization may hinder holistic understanding.}.


\begin{table}[t]
\centering
\begin{tabular}{lccccc}
\toprule
\multicolumn{1}{c}{\multirow{4}{*}{Method}} & \multicolumn{4}{c}{Question types (\#Question)}                    \\ \cmidrule(lr){2-5}
\multicolumn{1}{c}{}                        & \textit{Compo.} & \textit{Conj.} & \textit{Sup.} & \textit{Compa.} \\
\multicolumn{1}{c}{}                        & (1546)    & (1575)    & (197)     & (213)     \\ \midrule
\multicolumn{1}{c|}{}                        & \multicolumn{4}{c}{Gemma2-9b}                      \\ \midrule
\multicolumn{1}{c|}{ProgRAG}              & 70.8        &\textbf{76.6}&\textbf{70.6}&\textbf{75.6}\\
\multicolumn{1}{c|}{w/o \textit{P.E.} }       &\textbf{72.2}& 62.5        & 29.4        & 46.0        \\
\multicolumn{1}{c|}{w/o \textit{P.R.} }    & \underline{71.7}        & \underline{69.2}        & \underline{63.5}        & \underline{72.3}        \\ \midrule
\multicolumn{1}{c|}{}                        & \multicolumn{4}{c}{GPT-4o-mini}                       \\ \midrule
\multicolumn{1}{c|}{MFC}                     & 66.0        & 62.8        & 39.6        & 55.0        \\
\multicolumn{1}{c|}{PoG}                     & 67.4        & 60.2        & 36.0        & 57.3        \\
\multicolumn{1}{c|}{ToG}                     & 62.4        & 57.8        & 35.0        & 49.3        \\
\multicolumn{1}{c|}{SubgraphRAG}             & 61.4        & 66.9        & 39.1        & 51.2        \\ \bottomrule
\end{tabular}
\caption{Performance analysis for different question types. \textit{Compo.}, \textit{Conj.}, \textit{Sup.}, and \textit{Compa.} denote Composition, Conjunction, Superlative, and Comparative, respectively. w/o \textit{P.E.} and w/o \textit{P.R.} stand for without Prefix Enumeration and without Prefix Repacking, respectively.}
\label{tab:question_types}
\end{table}

\begin{figure}[t]
\centering
\includegraphics[width=\columnwidth]{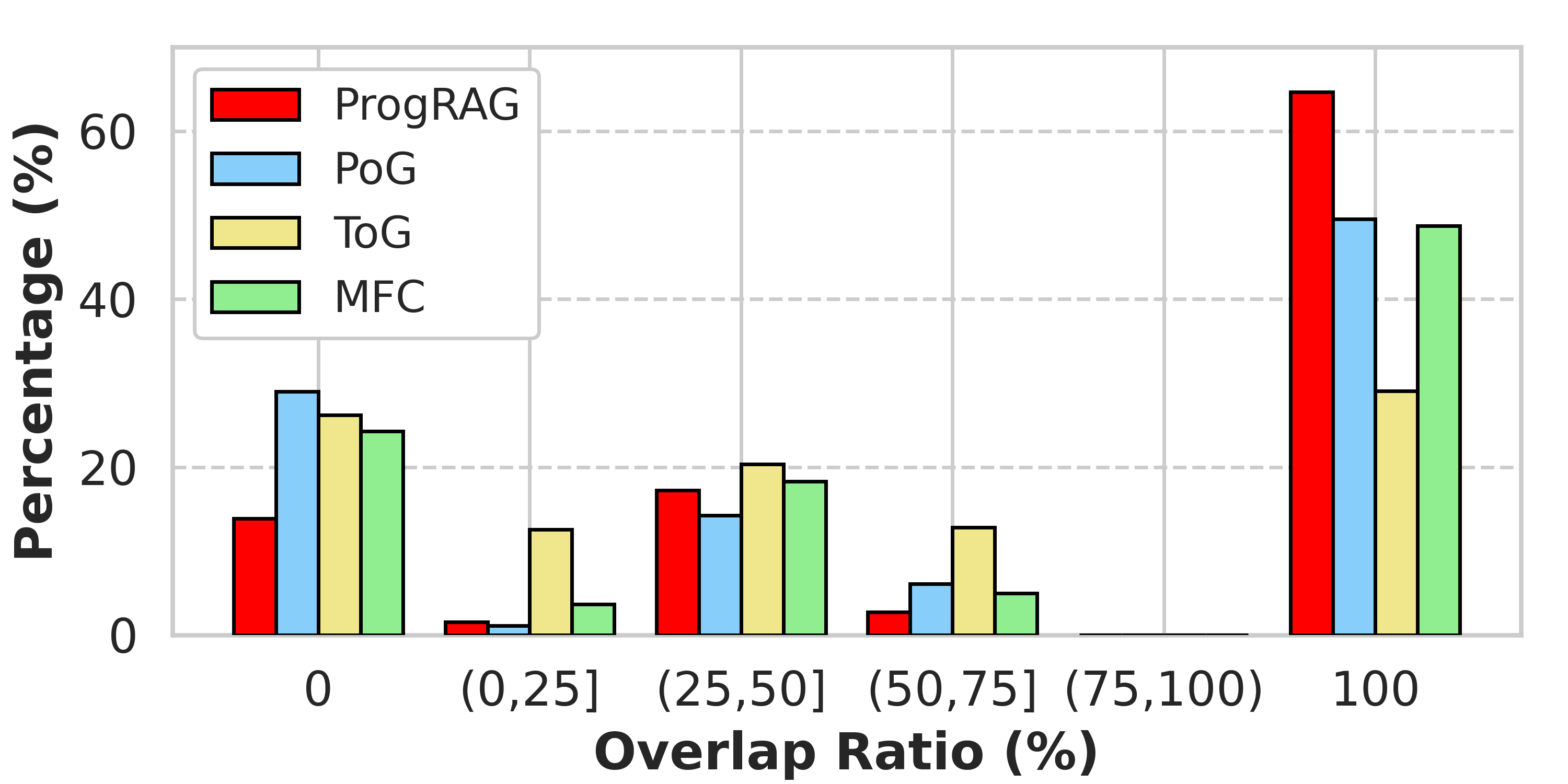}
\caption{Explored path overlap ratio on CWQ.}
\label{relover}
\end{figure}

\subsection{RQ3: Evaluation of Reasoning Path Retrieval}
To address RQ3, the retrieval performance of ProgRAG is evaluated along two key dimensions. First, we assess how effectively the retrieved paths cover the correct answer entities. ProgRAG achieves 101.5\% higher accuracy than ToG and 42\% higher than PoG, with detailed results in Appendix~E.
Second, we conduct a relation-level comparison between the reasoning paths retrieved by each method and the relations in ground-truth SPARQL queries to evaluate their structural alignment. Figure~\ref{relover} compares the overlap ratio \cite{sun2023think},i.e., the proportion of overlapping relations to the total number of relations in the ground-truth SPARQL path, for ours and representative iterative retrieve-and-reason methods such as PoG, ToG, and MFC.
ProgRAG achieves full relation overlap with ground-truth SPARQL queries in 65\% of questions and only 13\% with no overlap, outperforming all baselines. These results demonstrate the effectiveness of our progressive retrieval strategy, where relations are first retrieved by an external retriever and then pruned by the LLM at each iteration, yielding concise, semantically grounded reasoning paths.

\subsection{RQ4: Efficiency Analysis}

\begin{table}[t]
\setlength{\tabcolsep}{1.6mm}{
\centering
\begin{tabular}{lccccc}
\toprule
\multicolumn{1}{c}{Method} & \# Call & Time (s) & \# Path & \# Token & F1 \\ \midrule
ToG        & 27.7 & 72.4 & 5.9  & 899 & 41.9 \\
PoG        & 22.1 & 41.1 & 14.9 & 647 & 44.8 \\
MFC        & 17.6  & 43.7 & 3.3  & 685 & - \\
ProgRAG & 9.0  & 26.1 & 8.3 (3.9)  & 312 & 53.3 \\ \bottomrule
\end{tabular}
}
\caption{Efficiency analysis on CWQ. For ProgRAG, parentheses under \# Path show the average number of paths before prefix enumeration.}
\label{tab:cwq_effi}
\end{table}

Table~\ref{tab:cwq_effi} compares the efficiency of ProgRAG and the state-of-the-art KG-enhanced LLM inference methods, i.e., ToG, PoG, and MFC, on CWQ. All methods are evaluated by using GPT-4o-mini to ensure fair comparison. The table reports the average number of LLM calls, query time, the number of reasoning paths, the number of input tokens during reasoning, and F1 score. The retrieve-then-prune strategy of ProgRAG reduces redundant LLM calls and shortens input length, leading to consistent performance gains over all baselines. 
MFC uses fewer paths via abstracting mid-path nodes into meta entities but still produces long inputs due to coarse retrieval. Overall, more compact inputs consistently correlate with better performance, as illustrated in Figure~\ref{fig2}. 

\section{Conclusion}
In this paper, we introduce ProgRAG, a novel progressive retrieval and reasoning framework for multi-hop KGQA. ProgRAG decomposes a complex question into sub-questions, and incrementally constructs reasoning paths by iteratively retrieving candidate evidence with external retrievers and refining it through LLM-based pruning with uncertainty estimation. This progressive strategy enables more accurate retrieval and reasoning by dynamically optimizing the input context and mitigating hallucinations. Extensive experiments on benchmark datasets demonstrate that ProgRAG consistently outperforms state-of-the-art  baselines, achieving improved reasoning accuracy and reliability.

\section{Acknowledgments}
M. Park, H. Yang, and H. Kim were supported by Institute of Information \& communications Technology Planning \& Evaluation (IITP) grant funded by the Korea government(MSIT) (No.RS-2020-II201373, Artificial Intelligence Graduate School Program(Hanyang University)). K. Park was supported by IITP grant funded by the Korea government (MSIT) 
(No. 2018-0-00551, Framework of Practical Algorithms for NP-hard Graph Problems).

\bibliography{aaai2026}

\begin{thebibliography}{50}
\providecommand{\natexlab}[1]{#1}

\bibitem[{Ao et~al.(2025)Ao, Yu, Wang, Deng, Guo, Pang, Wang, Chua, Zhang, and Cai}]{ao2025lightprof}
Ao, T.; Yu, Y.; Wang, Y.; Deng, Y.; Guo, Z.; Pang, L.; Wang, P.; Chua, T.-S.; Zhang, X.; and Cai, Z. 2025.
\newblock Lightprof: A lightweight reasoning framework for large language model on knowledge graph.
\newblock In \emph{Proceedings of the AAAI Conference on Artificial Intelligence}, volume~39, 23424--23432.

\bibitem[{Axelsson and Skantze(2023)}]{axelsson2023using}
Axelsson, A.; and Skantze, G. 2023.
\newblock Using large language models for zero-shot natural language generation from knowledge graphs.
\newblock \emph{arXiv preprint arXiv:2307.07312}.

\bibitem[{Baek, Aji, and Saffari(2023)}]{baek2023knowledge}
Baek, J.; Aji, A.~F.; and Saffari, A. 2023.
\newblock Knowledge-augmented language model prompting for zero-shot knowledge graph question answering.
\newblock \emph{arXiv preprint arXiv:2306.04136}.

\bibitem[{Besta et~al.(2024)Besta, Blach, Kubicek, Gerstenberger, Podstawski, Gianinazzi, Gajda, Lehmann, Niewiadomski, Nyczyk et~al.}]{besta2024graph}
Besta, M.; Blach, N.; Kubicek, A.; Gerstenberger, R.; Podstawski, M.; Gianinazzi, L.; Gajda, J.; Lehmann, T.; Niewiadomski, H.; Nyczyk, P.; et~al. 2024.
\newblock Graph of thoughts: Solving elaborate problems with large language models.
\newblock In \emph{Proceedings of the AAAI conference on artificial intelligence}, volume~38, 17682--17690.

\bibitem[{Bollacker et~al.(2008)Bollacker, Evans, Paritosh, Sturge, and Taylor}]{bollacker2008freebase}
Bollacker, K.; Evans, C.; Paritosh, P.; Sturge, T.; and Taylor, J. 2008.
\newblock Freebase: a collaboratively created graph database for structuring human knowledge.
\newblock In \emph{Proceedings of the 2008 ACM SIGMOD international conference on Management of data}, 1247--1250.

\bibitem[{Brown et~al.(2020)Brown, Mann, Ryder, Subbiah, Kaplan, Dhariwal, Neelakantan, Shyam, Sastry, Askell et~al.}]{brown2020language}
Brown, T.; Mann, B.; Ryder, N.; Subbiah, M.; Kaplan, J.~D.; Dhariwal, P.; Neelakantan, A.; Shyam, P.; Sastry, G.; Askell, A.; et~al. 2020.
\newblock Language models are few-shot learners.
\newblock \emph{Advances in neural information processing systems}, 33: 1877--1901.

\bibitem[{Chen et~al.(2024)Chen, Tong, Jin, Sun, Ye, and Xiong}]{chen2024plan}
Chen, L.; Tong, P.; Jin, Z.; Sun, Y.; Ye, J.; and Xiong, H. 2024.
\newblock Plan-on-graph: Self-correcting adaptive planning of large language model on knowledge graphs.
\newblock \emph{Advances in Neural Information Processing Systems}, 37: 37665--37691.

\bibitem[{Dhole(2025)}]{dhole2025retrieve}
Dhole, K.~D. 2025.
\newblock To retrieve or not to retrieve? uncertainty detection for dynamic retrieval augmented generation.
\newblock \emph{arXiv preprint arXiv:2501.09292}.

\bibitem[{Guo, Toroghi, and Sanner(2024)}]{guo2024cr}
Guo, W.; Toroghi, A.; and Sanner, S. 2024.
\newblock Cr-lt-kgqa: A knowledge graph question answering dataset requiring commonsense reasoning and long-tail knowledge.
\newblock \emph{arXiv preprint arXiv:2403.01395}.

\bibitem[{He et~al.(2024)He, Tian, Sun, Chawla, Laurent, LeCun, Bresson, and Hooi}]{he2024g}
He, X.; Tian, Y.; Sun, Y.; Chawla, N.; Laurent, T.; LeCun, Y.; Bresson, X.; and Hooi, B. 2024.
\newblock G-retriever: Retrieval-augmented generation for textual graph understanding and question answering.
\newblock \emph{Advances in Neural Information Processing Systems}, 37: 132876--132907.

\bibitem[{Holtzman et~al.(2019)Holtzman, Buys, Du, Forbes, and Choi}]{holtzman2019curious}
Holtzman, A.; Buys, J.; Du, L.; Forbes, M.; and Choi, Y. 2019.
\newblock The curious case of neural text degeneration.
\newblock \emph{arXiv preprint arXiv:1904.09751}.

\bibitem[{Ji et~al.(2023)Ji, Lee, Frieske, Yu, Su, Xu, Ishii, Bang, Madotto, and Fung}]{ji2023survey}
Ji, Z.; Lee, N.; Frieske, R.; Yu, T.; Su, D.; Xu, Y.; Ishii, E.; Bang, Y.~J.; Madotto, A.; and Fung, P. 2023.
\newblock Survey of hallucination in natural language generation.
\newblock \emph{ACM computing surveys}, 55(12): 1--38.

\bibitem[{Jiang et~al.(2023{\natexlab{a}})Jiang, Zhou, Dong, Ye, Zhao, and Wen}]{jiang2023structgpt}
Jiang, J.; Zhou, K.; Dong, Z.; Ye, K.; Zhao, W.~X.; and Wen, J.-R. 2023{\natexlab{a}}.
\newblock Structgpt: A general framework for large language model to reason over structured data.
\newblock \emph{arXiv preprint arXiv:2305.09645}.

\bibitem[{Jiang et~al.(2023{\natexlab{b}})Jiang, Zhou, Zhao, Li, and Wen}]{jiang2023reasoninglm}
Jiang, J.; Zhou, K.; Zhao, W.~X.; Li, Y.; and Wen, J.-R. 2023{\natexlab{b}}.
\newblock Reasoninglm: Enabling structural subgraph reasoning in pre-trained language models for question answering over knowledge graph.
\newblock \emph{arXiv preprint arXiv:2401.00158}.

\bibitem[{Jiang et~al.(2022)Jiang, Zhou, Zhao, and Wen}]{jiang2022unikgqa}
Jiang, J.; Zhou, K.; Zhao, W.~X.; and Wen, J.-R. 2022.
\newblock Unikgqa: Unified retrieval and reasoning for solving multi-hop question answering over knowledge graph.
\newblock \emph{arXiv preprint arXiv:2212.00959}.

\bibitem[{Li, Miao, and Li(2024)}]{li2024simple}
Li, M.; Miao, S.; and Li, P. 2024.
\newblock Simple is effective: The roles of graphs and large language models in knowledge-graph-based retrieval-augmented generation.
\newblock \emph{arXiv preprint arXiv:2410.20724}.

\bibitem[{Li et~al.(2023)Li, Zhang, Zhang, Long, Xie, and Zhang}]{li2023towards}
Li, Z.; Zhang, X.; Zhang, Y.; Long, D.; Xie, P.; and Zhang, M. 2023.
\newblock Towards general text embeddings with multi-stage contrastive learning.
\newblock \emph{arXiv preprint arXiv:2308.03281}.

\bibitem[{Liang and Gu(2025)}]{liang2025fast}
Liang, X.; and Gu, Z. 2025.
\newblock Fast think-on-graph: Wider, deeper and faster reasoning of large language model on knowledge graph.
\newblock In \emph{Proceedings of the AAAI Conference on Artificial Intelligence}, volume~39, 24558--24566.

\bibitem[{Liu et~al.(2025{\natexlab{a}})Liu, Zhang, Lin, Yang, Peng, and Yin}]{liu2025symagent}
Liu, B.; Zhang, J.; Lin, F.; Yang, C.; Peng, M.; and Yin, W. 2025{\natexlab{a}}.
\newblock Symagent: A neural-symbolic self-learning agent framework for complex reasoning over knowledge graphs.
\newblock In \emph{Proceedings of the ACM on Web Conference 2025}, 98--108.

\bibitem[{Liu et~al.(2024)Liu, Zhang, Li, and Yao}]{liu2024explore}
Liu, G.; Zhang, Y.; Li, Y.; and Yao, Q. 2024.
\newblock Explore then determine: A gnn-llm synergy framework for reasoning over knowledge graph.
\newblock \emph{arXiv e-prints}, arXiv--2406.

\bibitem[{Liu et~al.(2023)Liu, Lin, Hewitt, Paranjape, Bevilacqua, Petroni, and Liang}]{liu2023lost}
Liu, N.~F.; Lin, K.; Hewitt, J.; Paranjape, A.; Bevilacqua, M.; Petroni, F.; and Liang, P. 2023.
\newblock Lost in the middle: How language models use long contexts.
\newblock \emph{arXiv preprint arXiv:2307.03172}.

\bibitem[{Liu et~al.(2025{\natexlab{b}})Liu, Halder, Qi, Xiao, Pappas, Htut, John, Benajiba, and Roth}]{liu2025towards}
Liu, S.; Halder, K.; Qi, Z.; Xiao, W.; Pappas, N.; Htut, P.~M.; John, N.~A.; Benajiba, Y.; and Roth, D. 2025{\natexlab{b}}.
\newblock Towards long context hallucination detection.
\newblock \emph{arXiv preprint arXiv:2504.19457}.

\bibitem[{Long et~al.(2025)Long, Zhuang, Li, Yao, and Wang}]{long2025eperm}
Long, X.; Zhuang, L.; Li, A.; Yao, M.; and Wang, S. 2025.
\newblock Eperm: An evidence path enhanced reasoning model for knowledge graph question and answering.
\newblock In \emph{Proceedings of the AAAI Conference on Artificial Intelligence}, volume~39, 12282--12290.

\bibitem[{Luo et~al.(2023{\natexlab{a}})Luo, Tang, Peng, Guo, Zhang, Ma, Dong, Song, Lin, Zhu et~al.}]{luo2023chatkbqa}
Luo, H.; Tang, Z.; Peng, S.; Guo, Y.; Zhang, W.; Ma, C.; Dong, G.; Song, M.; Lin, W.; Zhu, Y.; et~al. 2023{\natexlab{a}}.
\newblock Chatkbqa: A generate-then-retrieve framework for knowledge base question answering with fine-tuned large language models.
\newblock \emph{arXiv preprint arXiv:2310.08975}.

\bibitem[{Luo et~al.(2023{\natexlab{b}})Luo, Li, Haffari, and Pan}]{luo2023reasoning}
Luo, L.; Li, Y.-F.; Haffari, G.; and Pan, S. 2023{\natexlab{b}}.
\newblock Reasoning on graphs: Faithful and interpretable large language model reasoning.
\newblock \emph{arXiv preprint arXiv:2310.01061}.

\bibitem[{Luo et~al.(2024)Luo, Zhao, Haffari, Li, Gong, and Pan}]{luo2024graph}
Luo, L.; Zhao, Z.; Haffari, G.; Li, Y.-F.; Gong, C.; and Pan, S. 2024.
\newblock Graph-constrained reasoning: Faithful reasoning on knowledge graphs with large language models.
\newblock \emph{arXiv preprint arXiv:2410.13080}.

\bibitem[{Luo et~al.(2025)Luo, Zhao, Haffari, Phung, Gong, and Pan}]{luo2025gfm}
Luo, L.; Zhao, Z.; Haffari, G.; Phung, D.; Gong, C.; and Pan, S. 2025.
\newblock GFM-RAG: graph foundation model for retrieval augmented generation.
\newblock \emph{arXiv preprint arXiv:2502.01113}.

\bibitem[{Ma et~al.(2025)Ma, Chen, Zhou, Wang, and Zhang}]{ma2025estimating}
Ma, H.; Chen, J.; Zhou, J.~T.; Wang, G.; and Zhang, C. 2025.
\newblock Estimating LLM Uncertainty with Evidence.
\newblock \emph{arXiv preprint arXiv:2502.00290}.

\bibitem[{Mavromatis and Karypis(2024)}]{mavromatis2024gnn}
Mavromatis, C.; and Karypis, G. 2024.
\newblock Gnn-rag: Graph neural retrieval for large language model reasoning.
\newblock \emph{arXiv preprint arXiv:2405.20139}.

\bibitem[{Oord, Li, and Vinyals(2018)}]{oord2018representation}
Oord, A. v.~d.; Li, Y.; and Vinyals, O. 2018.
\newblock Representation learning with contrastive predictive coding.
\newblock \emph{arXiv preprint arXiv:1807.03748}.

\bibitem[{Pan et~al.(2024)Pan, Luo, Wang, Chen, Wang, and Wu}]{pan2024unifying}
Pan, S.; Luo, L.; Wang, Y.; Chen, C.; Wang, J.; and Wu, X. 2024.
\newblock Unifying large language models and knowledge graphs: A roadmap.
\newblock \emph{IEEE Transactions on Knowledge and Data Engineering}, 36(7): 3580--3599.

\bibitem[{Reimers and Gurevych(2019)}]{reimers2019sentence}
Reimers, N.; and Gurevych, I. 2019.
\newblock Sentence-bert: Sentence embeddings using siamese bert-networks.
\newblock \emph{arXiv preprint arXiv:1908.10084}.

\bibitem[{Reimers and Gurevych(2020)}]{reimers2020curse}
Reimers, N.; and Gurevych, I. 2020.
\newblock The curse of dense low-dimensional information retrieval for large index sizes.
\newblock \emph{arXiv preprint arXiv:2012.14210}.

\bibitem[{Sensoy, Kaplan, and Kandemir(2018)}]{sensoy2018evidential}
Sensoy, M.; Kaplan, L.; and Kandemir, M. 2018.
\newblock Evidential deep learning to quantify classification uncertainty.
\newblock \emph{Advances in neural information processing systems}, 31.

\bibitem[{Song et~al.(2020)Song, Tan, Qin, Lu, and Liu}]{song2020mpnet}
Song, K.; Tan, X.; Qin, T.; Lu, J.; and Liu, T.-Y. 2020.
\newblock Mpnet: Masked and permuted pre-training for language understanding.
\newblock \emph{Advances in neural information processing systems}, 33: 16857--16867.

\bibitem[{Sun et~al.(2023)Sun, Xu, Tang, Wang, Lin, Gong, Ni, Shum, and Guo}]{sun2023think}
Sun, J.; Xu, C.; Tang, L.; Wang, S.; Lin, C.; Gong, Y.; Ni, L.~M.; Shum, H.-Y.; and Guo, J. 2023.
\newblock Think-on-graph: Deep and responsible reasoning of large language model on knowledge graph.
\newblock \emph{arXiv preprint arXiv:2307.07697}.

\bibitem[{Talmor and Berant(2018)}]{talmor2018web}
Talmor, A.; and Berant, J. 2018.
\newblock The web as a knowledge-base for answering complex questions.
\newblock \emph{arXiv preprint arXiv:1803.06643}.

\bibitem[{Vrande{\v{c}}i{\'c} and Kr{\"o}tzsch(2014)}]{vrandevcic2014wikidata}
Vrande{\v{c}}i{\'c}, D.; and Kr{\"o}tzsch, M. 2014.
\newblock Wikidata: a free collaborative knowledgebase.
\newblock \emph{Communications of the ACM}, 57(10): 78--85.

\bibitem[{Wang, Ren, and Leskovec(2021)}]{wang2021relational}
Wang, H.; Ren, H.; and Leskovec, J. 2021.
\newblock Relational message passing for knowledge graph completion.
\newblock In \emph{Proceedings of the 27th ACM SIGKDD conference on knowledge discovery \& data mining}, 1697--1707.

\bibitem[{Wang et~al.(2025)Wang, Lin, Guo, Shun, Li, and Zhu}]{wang2025reasoning}
Wang, S.; Lin, J.; Guo, X.; Shun, J.; Li, J.; and Zhu, Y. 2025.
\newblock Reasoning of large language models over knowledge graphs with super-relations.
\newblock \emph{arXiv preprint arXiv:2503.22166}.

\bibitem[{Wang et~al.(2022)Wang, Wei, Schuurmans, Le, Chi, Narang, Chowdhery, and Zhou}]{wang2022self}
Wang, X.; Wei, J.; Schuurmans, D.; Le, Q.; Chi, E.; Narang, S.; Chowdhery, A.; and Zhou, D. 2022.
\newblock Self-consistency improves chain of thought reasoning in language models.
\newblock \emph{arXiv preprint arXiv:2203.11171}.

\bibitem[{Wei et~al.(2022)Wei, Wang, Schuurmans, Bosma, Xia, Chi, Le, Zhou et~al.}]{wei2022chain}
Wei, J.; Wang, X.; Schuurmans, D.; Bosma, M.; Xia, F.; Chi, E.; Le, Q.~V.; Zhou, D.; et~al. 2022.
\newblock Chain-of-thought prompting elicits reasoning in large language models.
\newblock \emph{Advances in neural information processing systems}, 35: 24824--24837.

\bibitem[{Xu et~al.(2025)Xu, Li, Zhang, Lin, Zhu, Zheng, Wu, Zhao, Xu, and Chen}]{xu2025harnessing}
Xu, D.; Li, X.; Zhang, Z.; Lin, Z.; Zhu, Z.; Zheng, Z.; Wu, X.; Zhao, X.; Xu, T.; and Chen, E. 2025.
\newblock Harnessing large language models for knowledge graph question answering via adaptive multi-aspect retrieval-augmentation.
\newblock In \emph{Proceedings of the AAAI Conference on Artificial Intelligence}, volume~39, 25570--25578.

\bibitem[{Yao et~al.(2023)Yao, Yu, Zhao, Shafran, Griffiths, Cao, and Narasimhan}]{yao2023tree}
Yao, S.; Yu, D.; Zhao, J.; Shafran, I.; Griffiths, T.; Cao, Y.; and Narasimhan, K. 2023.
\newblock Tree of thoughts: Deliberate problem solving with large language models.
\newblock \emph{Advances in neural information processing systems}, 36: 11809--11822.

\bibitem[{Yih et~al.(2016)Yih, Richardson, Meek, Chang, and Suh}]{yih2016value}
Yih, W.-t.; Richardson, M.; Meek, C.; Chang, M.-W.; and Suh, J. 2016.
\newblock The value of semantic parse labeling for knowledge base question answering.
\newblock In \emph{Proceedings of the 54th Annual Meeting of the Association for Computational Linguistics (Volume 2: Short Papers)}, 201--206.

\bibitem[{Yixing et~al.(2024)Yixing, Wang, Zhang, Liu, and Mao}]{yixing2024chain}
Yixing, P.; Wang, Q.; Zhang, L.; Liu, Y.; and Mao, Z. 2024.
\newblock Chain-of-question: A progressive question decomposition approach for complex knowledge base question answering.
\newblock In \emph{Findings of the Association for Computational Linguistics ACL 2024}, 4763--4776.

\bibitem[{Yu et~al.(2022)Yu, Zhang, Ng, Zhu, Li, Wang, Hu, Wang, Wang, and Xiang}]{yu2022decaf}
Yu, D.; Zhang, S.; Ng, P.; Zhu, H.; Li, A.~H.; Wang, J.; Hu, Y.; Wang, W.; Wang, Z.; and Xiang, B. 2022.
\newblock Decaf: Joint decoding of answers and logical forms for question answering over knowledge bases.
\newblock \emph{arXiv preprint arXiv:2210.00063}.

\bibitem[{Zhang et~al.(2025)Zhang, Zhu, Li, Yu, Kong, Wang, Miao, Zhang, and Zhou}]{zhang2025good}
Zhang, B.; Zhu, J.; Li, C.; Yu, H.; Kong, L.; Wang, Z.; Miao, D.; Zhang, X.; and Zhou, J. 2025.
\newblock What is a Good Question? Assessing Question Quality via Meta-Fact Checking.
\newblock In \emph{Proceedings of the AAAI Conference on Artificial Intelligence}, volume~39, 15248--15256.

\bibitem[{Zhao et~al.(2024)Zhao, Zhao, Wang, Wang, and Xu}]{zhao2024kg}
Zhao, R.; Zhao, F.; Wang, L.; Wang, X.; and Xu, G. 2024.
\newblock Kg-cot: Chain-of-thought prompting of large language models over knowledge graphs for knowledge-aware question answering.
\newblock In \emph{Proceedings of the Thirty-Third International Joint Conference on Artificial Intelligence (IJCAI-24)}, 6642--6650. International Joint Conferences on Artificial Intelligence.

\bibitem[{Zhu et~al.(2025)Zhu, Liu, Aizawa, and Shimodaira}]{zhu2025beyond}
Zhu, Y.; Liu, Q.; Aizawa, A.; and Shimodaira, H. 2025.
\newblock Beyond Chains: Bridging Large Language Models and Knowledge Bases in Complex Question Answering.
\newblock \emph{arXiv preprint arXiv:2505.14099}.

\end{thebibliography}

\clearpage

\appendix
\section*{Appendix}
\setcounter{secnumdepth}{1}
\section{Datasets}
We adopt three publicly available and challenging multi-hop KGQA datasets: WebQuestionsSP (WebQSP) \citep{yih2016value}, ComplexWebQuestions (CWQ) \citep{talmor2018web}, and CR-LT-KGQA \citep{guo2024cr}. WebQSP and CWQ are constructed based on Freebase \citep{bollacker2008freebase} and CR-LT-KGQA (CR-LT) is built on Wikidata \citep{vrandevcic2014wikidata}. Unlike Freebase, Wikidata lacks a strict relational hierarchy, leading to a more diverse and less structured relation space. In addition, CR-LT-KGQA includes questions about obscure or long-tail entities rarely seen in the training corpora of large language models, offering a more realistic evaluation setting that highlights the importance of external knowledge graphs when LLMs encounter entities beyond their internal knowledge. WebQSP comprises 2,826 training and 1,628 test instances with up to 2-hop reasoning, while CWQ includes 27,639 training and 3,531 test instances with up to 4-hop reasoning. CR-LT comprises 200 samples. For training the GNN used in our triple retrieval step, we utilized the question-specific subgraphs pre-extracted by RoG \citep{luo2023reasoning}. Specifically, for each key entity, we construct a single subgraph by merging the subgraphs of all training questions containing that entity. During inference, the subgraph corresponding to the key entity identified in the given question is used.

\section{Implementation Details}
We conduct all experiments with PyTorch on two NVIDIA A6000 GPUs. Unless noted otherwise, hyperparameters are consistent across datasets. For the relation retrieval of our method, we set $m=15$; for relation pruning, $n=3$. Triple retrieval uses top-$p$ sampling with $p$ as 0.9 and a temperature of 0.07 during the fine-tuning of MPNet. The GNN for our structure-based entity scoring uses $L=3$ layers for WebQSP and $L=6$ for CWQ. For uncertainty quantification, we apply an AU threshold of 1.55 with $l=4$, based on validation performance.

\section{Preliminary}
\textbf{Evidential modeling.} Evidential modeling for uncertainty quantification represents class probabilities as a Dirichlet distribution parameterized by evidence values predicted by the model, indicating the amount of support for each class. The total evidence controls the confidence of a model, with higher evidence—arising when the model assigns strong evidence to one or a few classes—producing a sharper Dirichlet distribution and thus lower uncertainty. For more details, see \citep{sensoy2018evidential}

\section{Definition of Variants for Ablation Study}\label{appendix: ablation_details}
To examine the individual contribution of each component in ProgRAG, we define the following ablated variants:
\begin{itemize}
\item \textbf{w/o Relation Retrieval}: Instead of retrieving relevant relations, this variant uses all 1-hop relations of each key entity as input for the relation pruning stage.
\item \textbf{w/o Relation Pruning}: This variant skips our relation pruning and directly uses the top-3 relations from the ranked candidate relation set $R_{re}(q_i)$ obtained from our triple retrieval.
\item \textbf{w/o Triple Retrieval}: All candidate triples in $T_{q_i}$ are used as input to the LLM without using external retrieval models.
\item \textbf{w/o Triple Pruning}: The retrieved top-$p$ triples are directly included in the partial reasoning paths without using any refinement in our triple pruning and uncertainty quantification.
\item \textbf{w/o Uncertainty Quantification}: In this variant, the triple pruning stage relies solely on the predicted answers from the LLM without performing uncertainty quantification.

 \item \textbf{w/o Prefix Enumeration}: 
 Only complete reasoning paths are provided as input to the LLM without enumerating their prefixes.
\item \textbf{w/o Prefix Repacking}: All prefixes obtained by prefix enumeration are randomly shuffled before being passed to the LLM, preventing semantic relevance-based ordering from guiding the reasoning process.
\item \textbf{w/o Question Decomposition}: Our retrieval and pruning techniques are performed on the original question without decomposition into sub-questions. To determine the exploration depth, we adopt a self-assessment prompting strategy commonly used in the iterative retrieve-and-reasoning approach \citep{sun2023think, chen2024plan, zhang2025good}, i.e., after each hop, the LLM is prompted to assess whether the currently explored path sufficiently answers the original question. We terminate if the LLM responds “Yes.” or the number of hops reaches the predefined maximum depth; proceed to the next hop otherwise.
\item \textbf{w/o Key Entity Mapping}: Following the Chain-of-Question method \citep{yixing2024chain}, the question is decomposed into multiple sub-questions without associating them with their corresponding key entities.
\end{itemize}

\section{Evaluation of Reasoning Path Retrieval}\label{entity_retrieval}


\begin{table}[t]
\resizebox{\columnwidth}{!}{
\centering
\begin{tabular}{lcccc}
\toprule
Method                & entity recall & avg. \# entity & entity hit \\ \midrule
ProgRAG (Gemma2-9b)           & 81.0          & 7.0            & 85.3       \\ 
PoG (GPT-4o-mini)             & 40.2          & 23.0           & 45.9       \\ 
ToG (GPT-4o-mini)             & 57.0          & 20.0           & 72.2       \\ \bottomrule
\end{tabular}
}
\caption{Performance of tail entity}
\label{tab:terminal}
\end{table}

Table \ref{tab:terminal} presents a comparison of reasoning path retrieval performance from an entity-level perspective on the CWQ dataset, evaluating ProgRAG against ToG and PoG. Specifically, \textit{entity recall} refers to the proportion of all answer entities that are included in the retrieved reasoning paths; \textit{entity hit} indicates the percentage of questions for which at least one correct answer entity appears in the reasoning path; \textit{avg. \# entity} denotes the average number of unique entities included in the reasoning paths. ProgRAG achieves significantly higher accuracy, reaching 81.0\% entity recall and 85.3\% entity hit with only an average of 7 compact entities. In contrast, ToG and PoG show much lower performance (57.0\% and 72.2\% for entity recall and hit in ToG, and 40.2\% and 45.9\% in PoG) even though ToG includes an average of 20 entities and PoG includes an average of 23 entities in their reasoning paths. This performance gap is primarily attributed to differences in retrieval strategies. ToG employs a fixed exploration width in the knowledge graph, which results in low recall for questions with a lot of answer entities. PoG adjusts the path width adaptively, but unlike ProgRAG, it lacks a dedicated pruning step to filter the retrieved information. This leads to the accumulation of excessive and potentially irrelevant information, increasing the likelihood of hallucinations by the LLM. Consequently, PoG often makes incorrect termination decisions at each iteration, prematurely halting the reasoning process. In contrast, ProgRAG constructs precise reasoning paths by performing triple retrieval conditioned on sub-questions and applying LLM-based pruning. This approach ensures that only semantically relevant triples are selected at each iteration, enabling more accurate and efficient multi-hop reasoning.

\section{Efficiency Analysis on WebQSP}
Table~\ref{tab:webq_effi} presents the efficiency analysis on the WebQSP dataset. ProgRAG consistently outperforms all baselines in both efficiency and effectiveness. It requires 42\% fewer LLM calls and is approximately 4.6 times faster than PoG. During the reasoning stage, ProgRAG uses 80\% fewer input tokens than ToG and 72\% fewer than PoG. Despite the reduced computational overhead, ProgRAG achieves significantly higher performance, with an F1 score of 75.9.

\begin{table}[t]
\resizebox{\columnwidth}{!}{
\centering
\renewcommand{\arraystretch}{1.0}
\begin{tabular}{lccccc}
\toprule
\multicolumn{1}{c}{Method} & \# Call & Time (s) & \# Path & \# Token & F1 \\ \midrule
ToG        & 12.8 & 50.9 & 3.6  & 850 & -    \\
PoG        & 8.8  & 35.9 & 8.8  & 620 & 59.4 \\
MFC        & 12.2  & 27.7 & 2.1  & 638 & - \\
ProgRAG & 5.1  & 7.8  & 3.9  & 171 & 75.9 \\ \bottomrule
\end{tabular}
}
\caption{Efficiency analysis on the WebQSP dataset.}
\label{tab:webq_effi}
\end{table}

\section{Uncertainty Quantification}

\begin{figure}[t]
\centering
\includegraphics[width=\columnwidth]{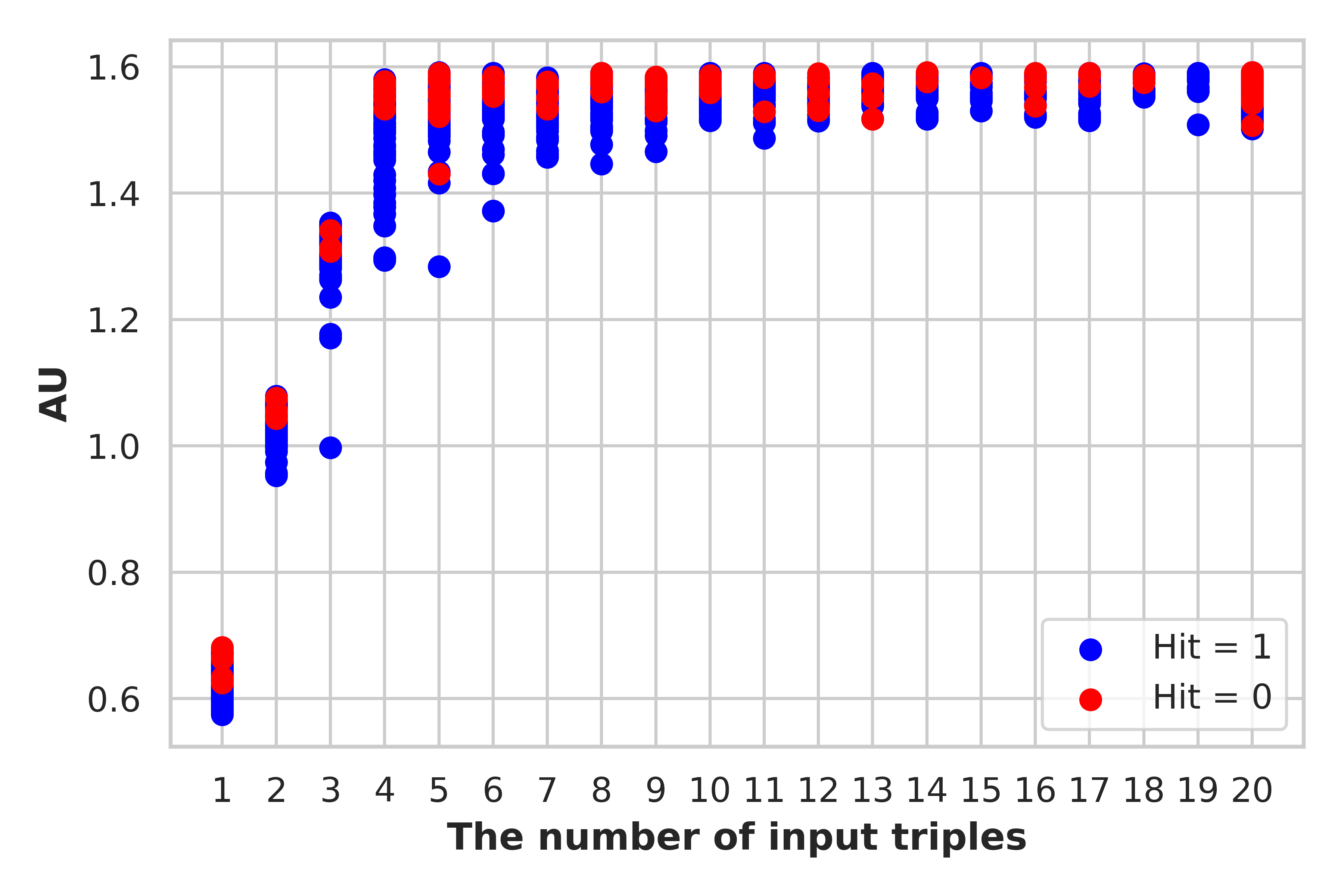}
\caption{Uncertainty trends by triple input size on the WebQSP dataset.}
\label{au}
\end{figure}

Figure \ref{au} shows the relationship between the number of input triples fed into the our triple pruning component and the uncertainty (i.e., AU) of the LLM, categorized by whether the triple pruner successfully returned a triple containing the correct answer entity. We conduct experiments on the WebQSP dataset and use \textit{Hit} metric to determine whether the output of the triple pruner includes any triples that contain the ground-truth answer entity: Hit = 1 if the triple pruner returns such a triple; Hit = 0 otherwise. As the number of input triples increases, the average AU also increases and eventually stabilizes, indicating that exposure to more candidate triples introduces more ambiguity. For each number of input triples, AU is generally higher in failure cases where Hit = 0 than in success cases where Hit = 1. These results demonstrate that our uncertainty quantification component effectively captures the reliability of LLM responses during the triple pruning step.

\begin{table}[t]
\centering
\begin{tabular}{lcccc}
\toprule
Method                    & 1 hop & 2 hop & $\geq 3$ hop \\ \midrule
ProgRAG (Gemma2-9b) & \textbf{71.6}  & \textbf{77.8}  & \textbf{66.6}  \\
PoG (GPT-4o-mini)         & 60.5  & 62.5  & 62.8  \\
ToG (GPT-4o-mini)         & 57.3  & 59.5  & 55.8  \\ \bottomrule
\end{tabular}
\caption{Performance analysis based on the number of relational steps on the CWQ dataset.}
\label{tab:hops}
\end{table}

\begin{figure}[t]
\centering
\includegraphics[width=\columnwidth]{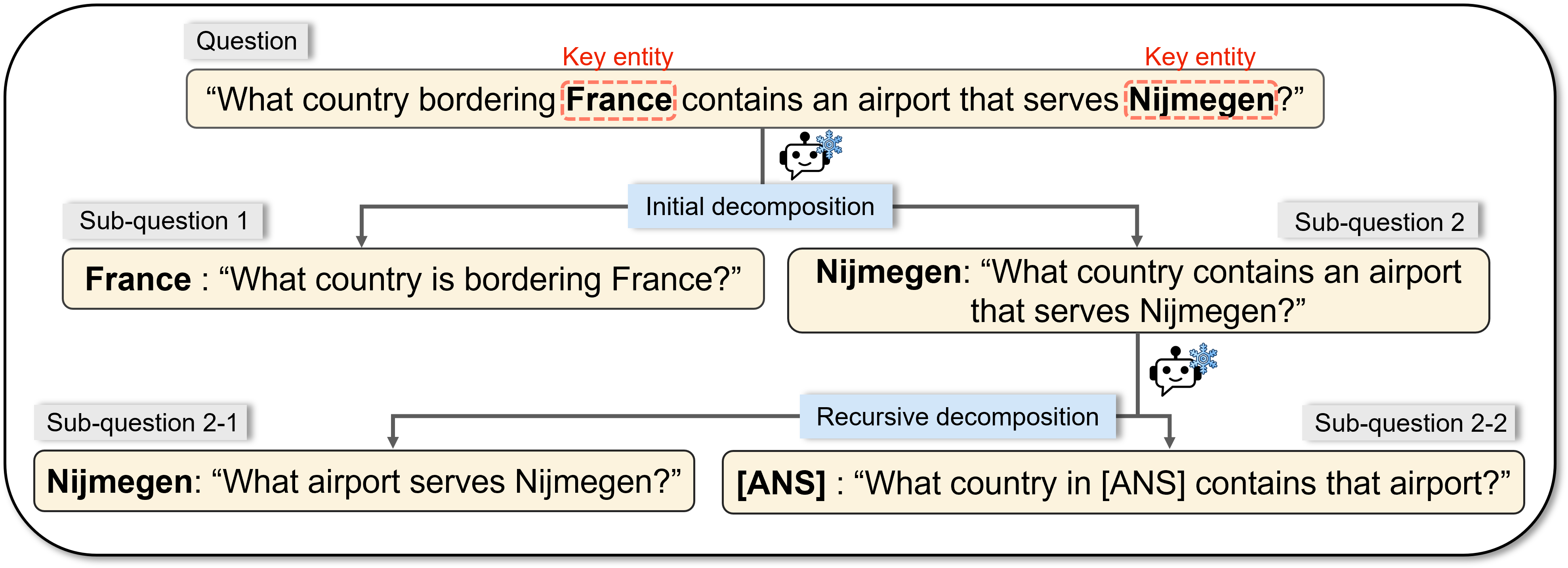}
\caption{An example of our question decomposition method.}
\label{fig_QD}
\end{figure}

\section{Performance Analysis by Number of Relational Steps}
Table \ref{tab:hops} presents an analysis of the performance of ProgRAG and the two KG-enhanced LLMs using the two iterative retrieval-and-reasoning framework (i.e., PoG and ToG) on the CWQ dataset, across different number of relational steps. Here, the number of relational steps is defined as the number of relational hops needed to traverse the KG to find the answer from the key entity. This dataset consists of 1-hop (28.0\%), 2-hop (65.9\%), and 3-hop or more (6.1\%) questions, with the majority requiring multi-hop reasoning. ProgRAG achieves the highest Hit@1 across all relational hops. It reaches 77.8 on 2-hop questions, representing a relative improvement of 24.5\% over PoG and 30.7\% over ToG. For questions with more than 2-hops, ProgRAG maintains stronger performance than the baselines, demonstrating robust stability under deeper reasoning demands. These results validate the effectiveness of our progressive retrieval and reasoning strategy, highlighting its strength for solving complex multi-hop reasoning tasks.

\section{Illustration of Question Decomposition}
Figure~\ref{fig_QD} provides an illustrative example of the question decomposition process. Given the question \textit{``What country bordering France contains an airport that serves Nijmegen?''}, the LLM identifies two key entities, i.e., \textit{France} and \textit{Nijmegen}, and decomposes the question into two sub-question chains.  
\begin{itemize}
    \item For \textit{France}: \textit{``What country borders France?''}  
    \item For \textit{Nijmegen}: \textit{``What airport serves Nijmegen?''} $\rightarrow$ \textit{``What country contains that airport?''}
\end{itemize}

This decomposition allows the model to explore distinct reasoning paths rooted in different key entities, facilitating structured and interpretable multi-hop reasoning.

\section{Ablation Study}

\begin{table}[t]
\centering
\renewcommand{\arraystretch}{1.0}
\begin{tabular}{lccc}
\toprule

Method    &  WebQSP        & CWQ \\ \midrule
ProgRAG                 & \textbf{88.5}          & \textbf{73.7}        \\
w/ MFC-based Decomposition  &  -          & 61.8        \\ 
\bottomrule
\end{tabular}
\caption{Experimental results comparing with existing question decomposition methods using Hit@1 as the evaluation metric.}
\label{tab:mfc}
\end{table}

We further replace our question decomposition method with that of MFC to assess the effectiveness of our approach (denoted as w/ MFC-based Decomposition). Table~\ref{tab:mfc} reports the performance, measured by Hit@1, when applying the question decomposition method adopted by MFC \citep{zhang2025good}. In contrast to our method that obtains in advance the full set of sub-questions, the MFC-based approach does not perform full decomposition upfront, resulting in a reasoning depth that is determined dynamically during inference. At each iteration, it decides whether to terminate based on the currently retrieved information. If not, it continues by prompting the LLM to decompose the next sub-question, using the original question, the current sub-question, and the retrieved information as inputs. In summary, while MFC  determines whether to decompose further at each step through LLM self-assessment, often leading to premature termination or unnecessary continuation, our method avoids such assessments. Instead, we decompose the question into key entity-specific sub-question chains, where the reasoning depth is defined by the number of sub-questions in each chain. This approach is more robust to LLM hallucinations and results in superior performance.


\section{Prompts}
Prompts used in our method are shown in Figures \ref{fig:qdecom_prompt} and \ref{fig:relprune_prompt}.

\section{Code}
The code is structured as follows:

\subsubsection{main.py} Main script for executing ProgRAG. You can execute the model using the command: \texttt{python3 main.py --dataset [Dataset\_Name]}.

\subsubsection{graph\_preprocess.py} This code constructs the partial KG used in our study, along with subgraphs corresponding to each key entity.

\subsubsection{GNN} Folders containing codes for structure-based triple retriever.

\subsubsection{GNN/train.py} It is used to train the GNN model employed as a triple retriever.

\subsubsection{MPNet} This folder contains the code for the Textual Semantic-based Triple Retriever. To fine-tune the retriever, run the corresponding bash script for each dataset (e.g., \texttt{webqsp.sh}).

\subsubsection{prompts.py} This file includes all prompts necessary for executing ProgRAG.

\begin{figure*}[t]
\centering
\begin{tcolorbox}[colback=gray!10, colframe=gray!80, title=Question Decomposition Prompt, width=\textwidth]
You are an expert in world knowledge with strong logical reasoning skills. \\ Your task is to identify the key entities and decompose the given question into sub-questions based on the key entities. Follow a step-by-step reasoning process, using each previous answer to guide the next step. \\
Each step must be logically and semantically connected to the previous one.\\
Sub-questions must be constructed using the words and phrases found in the original question. \\

Use the following tags when formatting your answer: \\
\text{[ANS]}: marks an answer (or intermediate result) that is used in subsequent steps, and typically represents a set of multiple candidate answers. \\

Case:\\
Decompose the question: ``What European Union country sharing borders with Germany contains the Lejre Municipality?" step by step: \\ 
The main entities are [``Germany", ``Lejre Municipality", ``Country"]. \\\\
Step 1:  
Generate the first sub-question: ``What European Union country shares borders with Germany?"  \\
The answer is [ANS1].\\\\
Step 2:  \\
Generate the second sub-question: ``What country contains the Lejre Municipality?"  \\
The answer is [ANS2].\\\\
Step 3:  \\
The final answer is [ANS3], which is the intersection of [ANS1] and [ANS2].\\
Here, [ANS1] and [ANS2] each represent a set of multiple candidate answers, and the final result [ANS3] includes only those entities that appear in both sets.\\

Return:\\
SUB-QUESTION1: What European Union country shares borders with Germany?\\
ENTITY1: Germany\\
SUB-QUESTION2: What country contains the Lejre Municipality? \\
ENTITY2: Lejre Municipality
\end{tcolorbox}
\caption{Prompt used for question decomposition.}
\label{fig:qdecom_prompt}
\end{figure*}

\begin{figure*}[t]
\centering
\begin{tcolorbox}[colback=gray!10, colframe=gray!80, title=Relation Pruning Prompt, width=\textwidth]
You are an expert of world knowledge with strong logical skills. \\
You have to retrieve the top 3 relations that are most relevant to the question from the candidate relations. \\
You must select the answer only from the given candidate relations.\\
If there is no relevant relation to return, then return "None". \\

Case: \\
Question: What sports team's owners are Jerry Jones? \\
Topic entity: Jerry Jones\\
Candidate relations: \text{[}sports.pro\_sports\_played.athlete, sports.pro\_athlete.teams, people.person.places\_lived, \\people.person.profession, common.topic.notable\_for, sports.sports\_team\_owner.teams\_owned, \\sports.pro\_athlete.sports\_played\_professionally, sports.sports\_team\_roster.player, people.person.employment\_history, \\sports.professional\_sports\_team.owner\_s\text{]}\\\\
Retrieve top 3 relations from question ``What sports team's owners are Jerry Jones?" step by step:\\\\
First:\\
We can assume that the answer is a team, as the question explicitly asks ``What sports team". \\
Therefore, we can infer that the answer type should be ``sports team" or ``team".\\\\
Second:\\
The question implies that ``Jerry Jones", the topic entity, owns a sports team.\\\\
Third:\\
Therefore, appropriate relations would describe the ownership of a sports team by a person.\\\\
Fourth:\\
Based on the candidate relations, the most relevant ones are\\ sports.sports\_team\_owner.teams\_owned, sports.professional\_sports\_team.owner\_s, sports.pro\_athlete.teams.\\\\

Return: sports.sports\_team\_owner.teams\_owned, sports.professional\_sports\_team.owner\_s, sports.pro\_athlete.teams
\end{tcolorbox}
\caption{Prompt used for relation pruning.}
\label{fig:relprune_prompt}
\end{figure*}

\end{document}